\documentclass[cameraready]{Interspeech}

\title{Cross-Attention is \underline{Half} Explanation in Speech-to-Text Models}

\author[affiliation={}, orcid=0000-0002-4494-8886]{Sara}{Papi}
\author[affiliation={}, orcid=0000-0002-0940-5595]{Dennis}{Fucci}
\author[affiliation={}, orcid=0000-0003-4217-1396]{Marco}{Gaido}
\author[affiliation={}, orcid=0000-0002-8811-4330]{Matteo}{Negri}
\author[affiliation={}, orcid=0000-0001-7480-2231]{Luisa}{Bentivogli}

\address{
    Fondazione Bruno Kessler, Italy
}

\email{\{spapi,mgaido,negri,bentivo\}@fbk.eu, dennis.fucci@gmail.com}


\usepackage{comment}
\usepackage{colortbl}
\usepackage[normalem]{ulem}
\useunder{\uline}{\ultab}{}
\usepackage{arydshln}
\usepackage{multirow}
\usepackage{graphicx}
\usepackage{soul}
\usepackage{amsmath}
\usepackage{mathabx}
\usepackage[most]{tcolorbox}
\usepackage{pifont}
\usepackage{csquotes}
\usepackage{subfigure}
\usepackage{wrapfig}
\usepackage[normalem]{ulem}
\definecolor{green}{HTML}{D0F4DE}
\definecolor{pink}{HTML}{FF99C8}
\definecolor{darkblue}{HTML}{02a6f7}
\definecolor{lightblue}{HTML}{A9DEF9}
\definecolor{purple}{HTML}{c064f5}
\definecolor{lightpurple}{HTML}{E4C1F9}
\definecolor{lightyellow}{HTML}{FDFCDC}
\definecolor{lightred}{HTML}{F28077}
\definecolor{darkcyan}{HTML}{1B96BA}
\definecolor{azure}{rgb}{0.0, 0.5, 1.0}
\definecolor{darkpurple}{HTML}{88185B}
\newcommand{\basem}[1]{\texttt{base}}
\newcommand{\smallm}[1]{\texttt{small}}
\newcommand{\largem}[1]{\texttt{large}}
\newcommand{\CA}[1]{\textcolor{purple}{\(\mathbf{CA}\)}}
\newcommand{\SM}[1]{\textcolor{darkblue}{\(\mathbf{SM}\)}}

\makeatletter
\newcommand\blfootnote[1]{%
  \begingroup
    \renewcommand\thefootnote{}%
    \let\orig@makefntext\@makefntext
    \def\@makefntext##1{\noindent##1}%
    \footnotetext{#1}%
    \addtocounter{footnote}{0}%
    \let\@makefntext\orig@makefntext
  \endgroup
}
\makeatother

\begin{document}

\maketitle


\begin{abstract}
    Cross-attention is widely used in speech-to-text (S2T) systems and often exploited for downstream applications such as timestamp prediction and speech-text alignment, under the assumption that it reflects input-output dependencies. While extensively debated in NLP, its explanatory role remains underexplored in the speech domain. We empirically assess the explanatory power of cross-attention in S2T models by comparing attention scores with input saliency maps from feature-attribution methods. Our analysis spans monolingual and multilingual, single-task and multi-task models at multiple scales. We find moderate alignment between attention and saliency, particularly when aggregating across heads and layers. However, cross-attention captures only about 50\% of input relevance and, at best, 52-75\% of the encoder saliency. These results show that cross-attention offers useful but incomplete explanatory cues and should be interpreted with caution as a proxy for model behavior in S2T systems.
\end{abstract}

\section{Introduction}
\label{sec:intro}
Cross-attention \cite{bahdanau2015neural} is the core mechanism of the encoder-decoder Transformer architecture \cite{NIPS2017_3f5ee243}, a model that has become foundational across numerous AI domains \cite{9194070,9859720,lee2023cast,xattn-medicine,Lu31122024}, including natural language processing (NLP). 
Designed for modeling dependencies between the generated output sequence and the input representations, the cross-attention scores--derived from the attention mechanism--have been leveraged in various NLP tasks \cite{hu2020introductory,10049971}, such as 
source-target textual alignment \cite{garg-etal-2019-jointly,chen-etal-2020-accurate}, co-reference resolution \cite{voita-etal-2018-context}, and word sense disambiguation \cite{tang-etal-2018-analysis}.

In speech-to-text (S2T) modeling, cross-attention scores have been widely repurposed for diverse downstream applications such as audio-text alignment \cite{zhao20j_interspeech,lee20e_interspeech}, speaker identification \cite{kim19b_interspeech}, timestamp estimation \cite{9747085,lintoai2023whispertimestamped,zusag24_interspeech}, and guiding simultaneous automatic speech recognition (ASR) and speech translation (ST) \cite{wang24ea_interspeech,papi-etal-2023-attention,papi23_interspeech}.
These applications rely on the implicit assumption that 
cross-attention reliably indicates what the model attends to in the input signal during output generation. However, despite its widespread use, this assumption has 
never been verified.
A key concern is that cross-attention operates over the encoder's output sequence--rather than directly on the raw audio--which may have been reorganized or mixed with contextual information. This phenomenon, known as context mixing \cite{mohebbi-etal-2023-quantifying}, can potentially obscure the alignment between cross-attention weights and the original input signal.
Similar concerns have been extensively debated in the NLP community, where the reliability of attention mechanisms as explanations has been both challenged and defended, leading to conflicting perspectives and empirical evidence \cite{serrano-smith-2019-attention,jain-wallace-2019-attention,wiegreffe-pinter-2019-attention,bastings-filippova-2020-elephant}. In contrast, this question remains largely underexplored in the speech domain. 
Existing work on explainability in S2T has primarily focused on self-attention \cite{shim2022understanding,audhkhasi22_interspeech,a-shams-etal-2024-uncovering}, or on empirically measuring the effects of context mixing \cite{mohebbi-etal-2023-homophone}, without directly assessing the explanatory potential of cross-attention mechanisms.

To address this gap, we present the first systematic 
analysis of cross-attention as a proxy for input-output dependencies in S2T models. 
Our study serves two main objectives: \textit{i)} assessing the validity of using cross-attention as a surrogate for input-output alignment,
and \textit{ii)} evaluating whether it provides insights comparable to formal explainability methods such as feature attribution--while being more lightweight and less computationally expensive to obtain \cite{samek-etal-2021, madsen-etal-2022}. 
We compare cross-attention scores with input saliency maps derived from SPES~\cite{fucci2025spes}, the current state-of-the-art feature-attribution method in S2T, to determine the extent to which
cross-attention 
captures 
which input features are relevant for models' predictions.
In addition, we compute saliency maps on
encoder outputs and compare them with cross-attention scores to evaluate whether cross-attention fully explains how the decoder uses encoded representations, avoiding potential 
context mixing effects.
Our 
analysis spans ASR and ST tasks across monolingual, multilingual, and multitask settings using 
state-of-the-art speech processing 
architectures \cite{gulati20_interspeech} at multiple scales. 
With consistent trends across different settings, we find 
that cross-attention exhibits moderate to strong correlations with input saliency maps and aligns more closely with encoder output representations, suggesting an influence of context mixing. However, the results also show that the overall explanatory power of cross-attention is limited--accounting for only $\sim$50\% of input relevance and, at best, 52-75\% of encoder output saliency.
Our findings uncover fundamental limitations in interpreting cross-attention as an explanatory proxy, suggesting that it provides an informative yet incomplete view of the factors driving predictions in S2T models.

\section{Related Works}
\textbf{Explainability in Speech-to-Text.}
Explainable AI (XAI) has emerged to make model behavior more interpretable to humans, thereby supporting informed decision-making and responsible deployment \cite{BARREDOARRIETA202082}.
%
While XAI research has seen a rapid growth in the last years across multiple modalities, including vision and language \cite{sharma2024exploring}, progress in the speech domain has lagged.
This gap arises from the inherent complexities of speech processing, including the multidimensional nature of speech signals across time and frequency, and the variability in output sequence length \cite{wu2024can}.
Despite these challenges, growing concerns about trustworthiness 
are driving explainability efforts in speech classification \cite{becker2024, pastor-etal-2024-explaining} and S2T generation \cite{wu2023explanations, wu2024can, fucci2025spes,mandel2016directly, kavaki2020identifying, trinh2020directly}. Most of these works rely on perturbation-based methods
that assess how input modifications affect model predictions \cite{covert-etal-2021, IVANOVS2021228}.
Among these, 
\cite{fucci2025spes} recently proposed 
a technique for autoregressive S2T models that identifies regions of the spectrogram that most influence predictions to generate saliency maps.
However, XAI methods are generally computationally expensive--especially perturbation-based approaches applied to large models \cite{luo2024understanding,202503.2318}--which motivates exploring whether cross-attention, already computed at inference time, could serve as a lightweight alternative in a landscape still lacking efficient explainability tools for speech models.

\noindent\textbf{Attention as Explanation.}
Attention mechanisms have been widely used to probe model behavior in text-based NLP, as attention scores often align with human intuitions about relevance and salience \cite{clark-etal-2019-bert,ferrando2024primer}.
Early studies proposed norm-based analyses to improve the interpretability of attention weights \cite{kobayashi-etal-2020-attention,kobayashi-etal-2021-incorporating,mohebbi-etal-2021-exploring,ferrando-etal-2022-measuring}, while others suggested aggregating attention across layers and heads to quantify input-output influence more systematically \cite{abnar-zuidema-2020-quantifying,ye-etal-2021-connecting, chefer2021transformer}.
While some 
raised concerns about whether attention reliably reflects which inputs are actually responsible for outputs \cite{jain-wallace-2019-attention,serrano-smith-2019-attention,bastings-filippova-2020-elephant}, others
proposed conditions under which attention can meaningfully explain model behavior \cite{wiegreffe-pinter-2019-attention}. More recent work highlights that attention aggregation may obscure localized, token-specific interactions \cite{modarressi-etal-2023-decompx,yang-etal-2023-local,oh-schuler-2023-token}, motivating hybrid approaches that combine attention with other XAI techniques, such as attribution methods \cite{modarressi-etal-2022-globenc}, or the use of attention as a regularization signal during interpretability-driven training \cite{xie-etal-2024-ivra}.
Despite the ongoing efforts, most research has focused on self-attention within encoders, with limited attention to feed-forward dynamics \cite{kobayashi2024analyzing} and even less to encoder-decoder models. A few studies have investigated attention in encoder-decoder architectures \cite{nguyen2021study}, including in machine translation \cite{zhou-etal-2024}, but cross-attention remains largely underexplored in the speech domain and 
absent from the broader \enquote{attention as explanation} debate in NLP. Our work seeks to bridge this gap by bringing cross-attention of S2T models into this broader conversation, aiming to assess
whether
it can serve as a reliable explanation--and where its limitations emerge.

\section{Methodology}
\label{sec:method}
\begin{figure}[ht]
    \centering
    \includegraphics[width=0.47\textwidth]{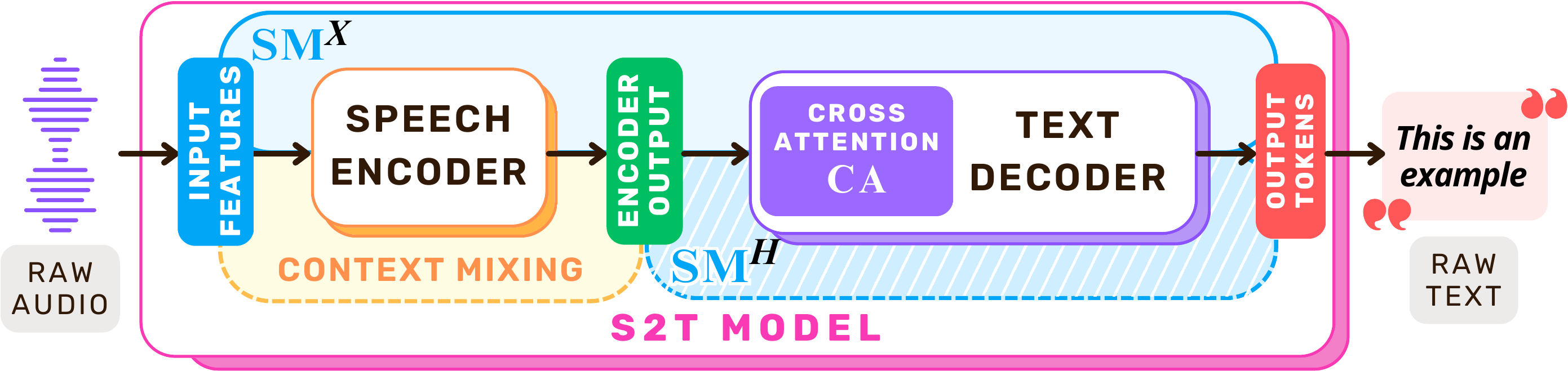}
    \caption{Visual representation of which part of the model is covered by input saliency maps \( \text{\SM{}}^X \) and encoder output saliency maps \( \text{\SM{}}^H \).}
    \label{fig:representation}
\end{figure}

We assess the extent to which cross-attention scores (\CA{}) explain how the model looks at
input features when generating a token
by comparing them to the saliency map on the input \( \text{\SM{}}^X \), obtained with the state-of-the art feature-attribution method for S2T, SPES \cite{fucci2025spes}. 
Additionally, to assess whether cross-attention more accurately reflects how the decoder accesses encoded representations--rather than capturing the model's full input-output behavior--we compare \CA{} with the encoder-output saliency map \( \text{\SM{}}^H \).
By analyzing how the correlation between \CA{} and \( \text{\SM{}}^H \) deviates from that with \( \text{\SM{}}^X \), 
we can indirectly quantify the impact of context mixing in the resulting explanations. 
A visual representation is presented in Figure \ref{fig:representation}.  
In the following, we first discuss how we extract 
\CA{}
scores (\S\ref{sec:xattn}), then how we compute 
\( \text{\SM{}}^X \) and \(\text{\SM{}}^H \) 
(\S \ref{subsec:xai_s2t}), and
how we compare them (\S \ref{sec:corr}).  

\subsection{Cross-Attention in Speech-to-Text}
\label{sec:xattn}

In S2T models, the cross-attention mechanism enables each 
output
token to integrate relevant portions of the encoded speech features, 
conditioning generation on the entire input.
Let \( \mathbf{X} \in \mathbb{R}^{T \times F} \) denote the speech input represented by mel-spectrogram features, where \( T \) is the number of time frames and \( F \) the number of frequency bins. The encoder processes \( \mathbf{X} \) into a sequence of hidden representations \( \mathbf{H} = \mathrm{Encoder}(\mathbf{X}) \in \mathbb{R}^{T' \times D} \), where \( T' < T \) reflects the number of encoder time steps after subsampling with a factor of $s$\footnote{As the length of the speech inputs is, in general, 10$\times$ longer that of the corresponding textual input, it is a common practice in S2T modeling to downsample the input through convolutional modules \cite{wang-etal-2020-fairseq}.} and \( D \) is the hidden dimensionality. The decoder then autoregressively generates an output sequence \( \mathbf{y} = (y_0, y_1, \ldots, y_I) \) of length $I$, where each token \( y_i \) is predicted based on the previously generated tokens \( y_{<i} \) and the encoder output \( \mathbf{H} \).
At each decoder layer \( \ell \in \{1, \ldots, L\} \), cross-attention scores are computed via dot-product attention \cite{graves2014neural} between the decoder's current hidden states \( \mathbf{B}^{(\ell)} \in \mathbb{R}^{I \times D} \) and the encoder outputs \( \mathbf{H} \). Specifically, the decoder states are linearly projected to queries \( \mathbf{Q}^{(\ell)} = \mathbf{B}^{(\ell)} \mathbf{W}_Q^{(\ell)} \in \mathbb{R}^{I \times d_k} \), while the encoder outputs are projected to keys \(
\mathbf{K}^{(\ell)} = \mathbf{H} \mathbf{W}_K^{(\ell)} \in \mathbb{R}^{T' \times d_k} \) 
using learned projection matrices \( \mathbf{W}_Q^{(\ell)}, \mathbf{W}_K^{(\ell)}\). The resulting cross-attention matrix \CA{} is:
\[
\text{\CA{}}^{(\ell)} = \mathrm{softmax} \left( \frac{\mathbf{Q}^{(\ell)} \mathbf{K}^{(\ell)\top}}{\sqrt{d_k}} \right) \in \mathbb{R}^{I \times T'} 
\]
where each row \( \text{\CA{}}^{(\ell)}_i \) 
represents the attention over encoder time steps for generating
token \( y_i \) at layer \( \ell \).
%
To capture diverse patterns, models employ multi-head attention \cite{NIPS2017_3f5ee243}, where each head \( h \in \{1, \ldots, H\} \) uses separate learned projections:
\[
\mathbf{Q}_h^{(\ell)} = \mathbf{B}^{(\ell)} \mathbf{W}_{Q,h}^{(\ell)}, \quad
\mathbf{K}_h^{(\ell)} = \mathbf{H} \mathbf{W}_{K,h}^{(\ell)}
\]
where \( \mathbf{W}_{Q,h}^{(\ell)} \in \mathbb{R}^{D \times d_k} \), and \( \mathbf{W}_{K,h}^{(\ell)} \in \mathbb{R}^{D \times d_k} \).
These projections are used to compute head-specific attention scores, yielding one attention matrix per head and layer: \( \{ \text{\CA{}}_h^{(\ell)} \}_{\ell=1, h=1}^{L,H} \).

Extracting  
the full set of scores provides a fine-grained view of how each output token in the generated hypothesis attends to the encoder's representations across all layers and heads.
To derive a single layer-wise or head-wise attention distribution, we compute the mean of the attention matrices over a subset \( \mathcal{S} \subseteq \{1, \dots, L\} \times \{1, \dots, H\} \) of layers and heads:
\[
\widebar{\text{\CA{}}}^{(\mathcal{S})} = \frac{1}{|\mathcal{S}|} \sum_{(\ell, h) \in \mathcal{S}} \text{\CA{}}_h^{(\ell)} \in \mathbb{R}^{I \times T'}
\]
By selecting different index sets \( \mathcal{S} \), this formulation yields layer-wise, head-wise, or global averages (e.g., setting \( \mathcal{S} = \{(\ell, h) : h = 1, \dots, H\} \) gives the average across heads at a given layer \( \ell \); \( \mathcal{S} = \{(\ell, h) : \ell = 1, \dots, L\} \) averages across layers for head \( h \); and \( \mathcal{S} = \{1, \dots, L\} \times \{1, \dots, H\} \) computes the full average).
These averages provide aggregated views of the model's attention patterns at a specific layer or head, summarizing how the model attends to the input speech over time.

\subsection{Feature Attribution for Speech-to-Text}
\label{subsec:xai_s2t}


To analyze how S2T models associate output tokens with regions of the speech input or internal representations (e.g., encoder outputs), we employ \textit{feature-attribution} methods that generate token-level \textit{saliency maps}, quantifying the relevance of different input segments to each prediction.

\noindent\textbf{Input Saliency Maps.}
Let 
\( \mathbf{X} \in \mathbb{R}^{T \times F} \) denote a mel-spectrogram input, where \( T \) is the number of time frames and \( F \) the number of frequency bins, and \( \mathbf{y} = (y_0, y_1, \ldots, y_I) \) the sequence of length \( I \) of the autoregressively-generated tokens predicted based on the input and the previously generated tokens  \( y_{<i} \). 
To attribute the prediction of each token \( y_i \) to specific parts of the input spectrogram, we adopt SPES~\cite{fucci2025spes}, 
the state-of-the-art feature-attribution method designed for autoregressive S2T modeling. SPES assigns a saliency score to each time-frequency 
element
of \( \mathbf{X} \), producing a saliency map \( \text{\SM{}}_i^X \in \mathbb{R}^{T \times F} \) for each token \( y_i \), where higher values indicate greater relevance of the corresponding time-frequency regions. SPES operates by clustering spectrogram elements based on energy profiles--capturing acoustic
components such as harmonics and background noise--and estimating the influence of each cluster by perturbing it with probability \( p_{X} \), repeated \( N_{X} \) times. The effect of each perturbation--i.e., masking parts of the input with 0 values--is measured by computing the Kullback-Leibler (KL) divergence \cite{kullback1951information} between the model's original output distribution \( P(y_i \mid y_{<i}, \mathbf{X}) \) and the distribution resulting from the perturbed input \( P^{(n)}(y_i \mid y_{<i}, \tilde{\mathbf{X}}^{(n)}) \) at time \( n \in \{1, \ldots, N_X\} \):
\[
\mathrm{KL}^{(n)}_i = \mathrm{KL} \left( P(y_i \mid y_{<i}, \mathbf{X}) \,\|\, P^{(n)}(y_i \mid y_{<i}, \tilde{\mathbf{X}}^{(n)}) \right)
\]
The divergence scores are then mapped back to the corresponding cluster positions in the spectrogram and aggregated to form the token-specific saliency map \( \text{\SM{}}_i^X \in \mathbb{R}^{T \times F} \). Stacking all saliency maps across the output 
\( \mathbf{y} \) yields a 3D saliency map:
\[
\text{\SM{}}^X = (\text{\SM{}}_0^X, \text{\SM{}}_1^X, \ldots, \text{\SM{}}_I^X) \in \mathbb{R}^{I \times T \times F}
\]
where each slice \( \text{\SM{}}_i^X[t, f] \) quantifies the contribution of the spectrogram bin at time \( t \) and frequency \( f \) to the 
token \( y_i \).

\noindent\textbf{Encoder Output Saliency Maps.} 
We further examine the influence of the encoder's internal representations on the prediction of each output token.
Let 
\( \mathbf{H} = \mathrm{Encoder}(\mathbf{X}) \in \mathbb{R}^{T' \times D} \) denote the sequence of encoder hidden states or \textit{encoder output}, where \( T' \) is the subsampled time dimension and \( D \) is the hidden dimension.
To assess the importance of the encoder output representations, we compute token-specific saliency maps \( \text{\SM{}}_i^{H} \in \mathbb{R}^{T' \times 1} \), where each entry reflects the contribution of the 
hidden state to the generation of \( y_i \).
Each encoder state \( \mathbf{H} = (\mathbf{h}_1, \ldots, \mathbf{h}_{T'}) \) is perturbed--setting all features to 0--independently
with probability \( p_H \), and the process is repeated \( N_H \) times. The KL divergence is computed for each perturbation between the original and perturbed output distributions:
\[
\mathrm{KL}^{(n)}_i = \mathrm{KL} \left( P(y_i \mid y_{<i}, \mathbf{H}) \,\|\, P^{(n)}(y_i \mid y_{<i}, \tilde{\mathbf{H}}^{(n)}) \right)
\]
The divergence scores are aggregated across perturbation trials to form the final saliency map, following the same strategy of SPES for the input-level saliency maps:
\[
\text{\SM{}}^H = (\text{\SM{}}_0^H, \text{\SM{}}_1^H, \ldots, \text{\SM{}}_I^H) \in \mathbb{R}^{I \times T'}
\]
where \(\text{\SM{}}^H\) captures the temporal relevance of the encoder's internal sequence representations for each output token \( y_i \).

\subsection{Correlation}
\label{sec:corr}


Since our focus lies in the temporal dynamics of the input \( \mathbf{X} \), we aggregate the 3D saliency scores \(\text{\SM{}}^X \in \mathbb{R}^{I \times T \times F}\) across the frequency dimension and downsample the time axis to produce a compressed representation \( \text{\SM{}}^X \in \mathbb{R}^{I \times T'} \) compatible with the cross-attention granularity, where \( T' \) corresponds to the number of encoder time steps. The aggregation is performed by taking the maximum saliency value over the frequency axis and within each corresponding time window. The resulting saliency map of each token \( \text{\SM{}}_i^X \in \mathbb{R}^{T' \times 1} \) reflects the temporal relevance of the input spectrogram with respect to the generation of token \( y_i \). Complementary experiments on the choice of the aggregation function are presented in \S\ref{subsec:aggregation-fun}.
%
%
%
%
Both \CA{} and \SM{} representations are normalized before computing the correlation scores, and the beginning and end of sentence are removed as they are not relevant for the analysis. 
The \CA{} matrix is normalized frame-wise using mean-variance normalization to mitigate the impact of potential attention sinks at initial or final tokens~\cite{clark-etal-2019-bert,ferrando-etal-2022-towards,papi-etal-2023-attention,xiao2024efficient} on the correlation computation.
Both \(\text{\SM{}}^X\) and \(\text{\SM{}}^H\) are normalized along the token dimension using the strategy proposed by \cite{fucci2025spes}, as saliency scores can vary widely across tokens due to differences in the original output distributions used to compute the KL divergence.

Following prior work on cross-attention analysis \cite{vig-belinkov-2019-analyzing} and explainable AI \cite{eberle-etal-2023-rather}, we use Pearson correlation to quantify the relationship between cross-attention scores and saliency-based explanations. Pearson is preferred over rank-based measures (Kendall, Spearman) because saliency scores are continuous and their magnitude--not just their ordering--is meaningful. In particular, rank-based correlations are highly sensitive to arbitrary ordering among non-informative features with near-zero scores, whereas Pearson better reflects agreement on whether a feature is important (high saliency) or not (low saliency).
Specifically, given 
\( \text{\CA{}}, \text{\SM{}} \in \mathbb{R}^{I \times T'} \), we compute the Pearson correlation coefficient $\rho$ to assess the similarity of their attribution patterns across output tokens and time steps. We first flatten each matrix into a vector of size \( I \cdot T' \):
\[
\mathbf{ca} = \mathrm{vec}(\text{\CA{}}), \quad \mathbf{sm} = \mathrm{vec}(\text{\SM{}}), \quad \mathbf{ca}, \mathbf{sm} \in \mathbb{R}^{I \cdot T'}
\]
The Pearson correlation coefficient \( \rho \in [-1, 1] \) is computed as:
\[
\rho(\text{\CA{}}, \text{\SM{}}) = \frac{\sum_{k=1}^{I \cdot T'} (\mathbf{ca}_k - \widebar{\mathbf{ca}})(\mathbf{sm}_k - \widebar{\mathbf{sm}})}{\sqrt{\sum_{k=1}^{I \cdot T'} (\mathbf{ca}_k - \widebar{\mathbf{ca}})^2} \sqrt{\sum_{k=1}^{I \cdot T'} (\mathbf{sm}_k - \widebar{\mathbf{sm}})^2}}
\]
where \( \widebar{\mathbf{ca}} \) and \( \widebar{\mathbf{sm}} \) denote the means of vectors \( \mathbf{ca} \) and \( \mathbf{sm} \):
\[
\widebar{\mathbf{ca}} = \frac{1}{I \cdot T'} \sum_{k=1}^{I \cdot T'} \mathbf{ca}_k, \quad \widebar{\mathbf{sm}} = \frac{1}{I \cdot T'} \sum_{k=1}^{I \cdot T'} \mathbf{sm}_k
\]
This scalar quantifies the linear relationship between the two saliency maps, with values closer to 1 indicating a strong positive correlation, and values near 0 indicating no correlation.

\section{Experimental Settings}
\label{sec:exp_settings}
\subsection{Models}

To avoid potential data contamination issues \cite{sainz-etal-2023-nlp}, we trained from scratch a monolingual ASR model and used two-sized multitask (ASR and ST) and multilingual (English and Italian) open-science models, FAMA \cite{papi-etal-2025-fama}, for which training data are completely open. 
All models are based on a Conformer encoder \cite{gulati20_interspeech} and a Transformer decoder, as Conformer is the current state-of-the-art architecture for S2T processing \cite{9414858,9746490,10023042}. 
The monolingual ASR model (\texttt{base}) is composed of 12 encoder layers and 6 decoder layers. Each layer has 8 attention heads, 512 as embedding dimension, and  FFNs dimension of 2,048. The vocabulary is built using a SentencePiece unigram model \cite{kudo-richardson-2018-sentencepiece} with size 8,000 trained on \textit{en} transcripts. The resulting number of parameters is 125M.
The multitask and multilingual FAMA models are of two sizes, \texttt{small} and \texttt{large}, the first having 12 encoder layers and 6 decoder layers and the latter having 24 encoder layers and 12 decoder layers. In both sizes, each layer has 16 attention heads, an embedding dimension of 1,024, and an FFN dimension of 4,096. The vocabulary is built using a SentencePiece unigram model with size 16,000 trained on \textit{en} and \textit{it} transcripts. Two extra tokens--\texttt{<lang:en>} and \texttt{<lang:it>}--indicate whether the target text is in \textit{en} or \textit{it}. The resulting number of parameters is 474M for the \texttt{small} model and 878M for the \texttt{large} model.
In all models, the Conformer encoder is preceded by two 1D convolutional layers with stride 2 and kernel size 5, resulting in a fixed subsampling factor $s$ of 4. The kernel size of the Conformer convolutional module is 31 for both the point- and depth-wise convolutions. The input audio is represented by 80 Mel-filterbank features extracted every 10 ms with a window of 25 ms.

All models were trained using a combination of three losses: \emph{i)} a label-smoothed cross-entropy loss
applied to the decoder output using the target text as the reference (transcripts for ASR and translations for ST), \emph{ii)} a CTC loss \cite{10.1145/1143844.1143891} computed using transcripts as reference 
on the output of the 8\textsuperscript{th} encoder layer for \texttt{base} and \texttt{small} and the 16\textsuperscript{th} for \texttt{medium}, \emph{iii)} a CTC loss on the final encoder output 
applied to predict the target text \cite{yan-etal-2023-ctc}. 
%
FAMA models 
followed a two-stage training procedure: ASR-only pre-training followed by joint ASR+ST training.
For ASR pre-training, the \texttt{small} model used the same schedule as the \texttt{base} model, while the \texttt{medium} model adopted a piecewise warm-up to avoid divergence \cite{peng24b_interspeech,gaido-etal-2025-warmup}.
During ASR+ST training, ASR and ST targets were sampled with equal probability, and lr was fixed at a constant value. 
All training were done using the FBK-fairseq repository \cite{papi-etal-2024-good}. 
%

\subsection{Data}
\label{subsec:data}

For the monolingual ASR model, we leverage the speech-to-text English data available for the \href{https://iwslt.org/2024/offline}{\texttt{\underline{IWSLT 2024}}} evaluation campaign, namely: CommonVoice \cite{ardila-etal-2020-common}, CoVoST v2 \cite{wang21s_interspeech}, Europarl-ST \cite{europarlst}, LibriSpeech \cite{librispeech}, MuST-C v1 \cite{di-gangi-etal-2019-must}, TEDLIUM v3 \cite{tedlium}, and VoxPopuli ASR \cite{wang-etal-2021-voxpopuli}. The resulting training set is about 3k hours of speech.
For the multitask (ASR and ST) multilingual large-scale models, more than 150k hours of open-source speech
in English (\textit{en}) and Italian (\textit{it}) were used: CommonVoice, CoVoST v2, FLEURS \cite{10023141}, MOSEL \cite{gaido-etal-2024-mosel}, MLS \cite{pratap20_interspeech}, and \href{https://hf.co/datasets/PleIAs/YouTube-Commons}{\texttt{YouTube-Commons}}.
For datasets missing the translations, they were generated using \texttt{MADLAD-400 3B-MT} \cite{kudugunta2023madlad}. This setting allows us to verify our analysis with a large-scale setting similar to the scale of a popular model, such as OWSM \cite{10389676} and 2 times that of NVIDIA Canary \cite{puvvada24_interspeech} while having complete control of data used during training, completely preventing data contamination issues.
Being the only non-synthetic dataset supporting both tasks and language directions, we select EuroParl-ST \cite{europarlst} as the test set for our analyses. The test set covers both \textit{en} and \textit{it} ASR, and \textit{en-it} and \textit{it-en} ST. 
The \textit{it/it-en} section consists of 1,686 segments, for a total of approximately 6 hours of audio, while the \textit{en/en-it} section contains 1,130 segments, for a total of approximately 3 hours of audio.

\subsection{Evaluation Process}

\textbf{Hypothesis and Cross-Attention Generation.}
Hypotheses are generated using beam search with beam 
5 and a no-repeat n-gram 
5. Attention scores are extracted from the selected layers or heads during decoding. ASR and ST quality metrics are reported in \S\ref{subsec:model_quality}.
Inference is performed on a single NVIDIA A40 GPU (40GB) with a batch size of 40k tokens, taking approximately 2.5 minutes for \basem{}, 3/5.5 minutes for \smallm{}, and 3/6.5 minutes for \largem{}, depending on the 
language.

\noindent\textbf{Explanation Heatmaps Generation.}
Following the best SPES configuration \cite{fucci2025spes}, we use the Morphological Fragmental Perturbation Pyramid \cite{9413046} for clustering, based on SLIC \cite{6205760}, a k-means-based algorithm that groups elements by spectral similarity.
Default parameters are adopted: a threshold length of 7.5 s, \href{https://scikit-image.org/docs/dev/api/skimage.segmentation.html\#skimage.segmentation.slic}{SLIC} sigma set to 0, compactness to 0.1, and MFPP patch rates of [400, 500, 600] per second.
We set $p_X=0.5$ and $N_X=20{,}000$ following \cite{fucci2025spes}. Input explanation quality is reported in \S\ref{subsec:model_quality}.
For hidden representations, we fix $N_H=20{,}000$ (i.e., the same number of iterations of $N_X$) and tune $p_H$ on the dev set, obtaining $p_H=0.7$.
Inference is performed using a single NVIDIA A40 GPU (40GB RAM) 
and takes $\sim$27 hours for \basem{}, $\sim$3-4 days for \smallm{} and $\sim$6-8 days for \largem{}, depending on the 
language.

\noindent\textbf{Correlation Computation.}
The Pearson $r$ correlation score is computed using the \texttt{scipy} implementation
and averaging across samples in the test set.

All outputs and the aggregation script are released under the CC-BY 4.0 and Apache 2.0 licenses, respectively, at: \href{https://fbk-my.sharepoint.com/:f:/g/personal/spapi_fbk_eu/IgC8uMU4Ig69QISVwFE-IMCeATH_l7rvQLftzkFr3Rc9XUY?e=hvl6w8}{\texttt{\underline{FBK SharePoint}}}.

\begin{figure}[!ht]
    \centering
    \includegraphics[width=0.45\textwidth]{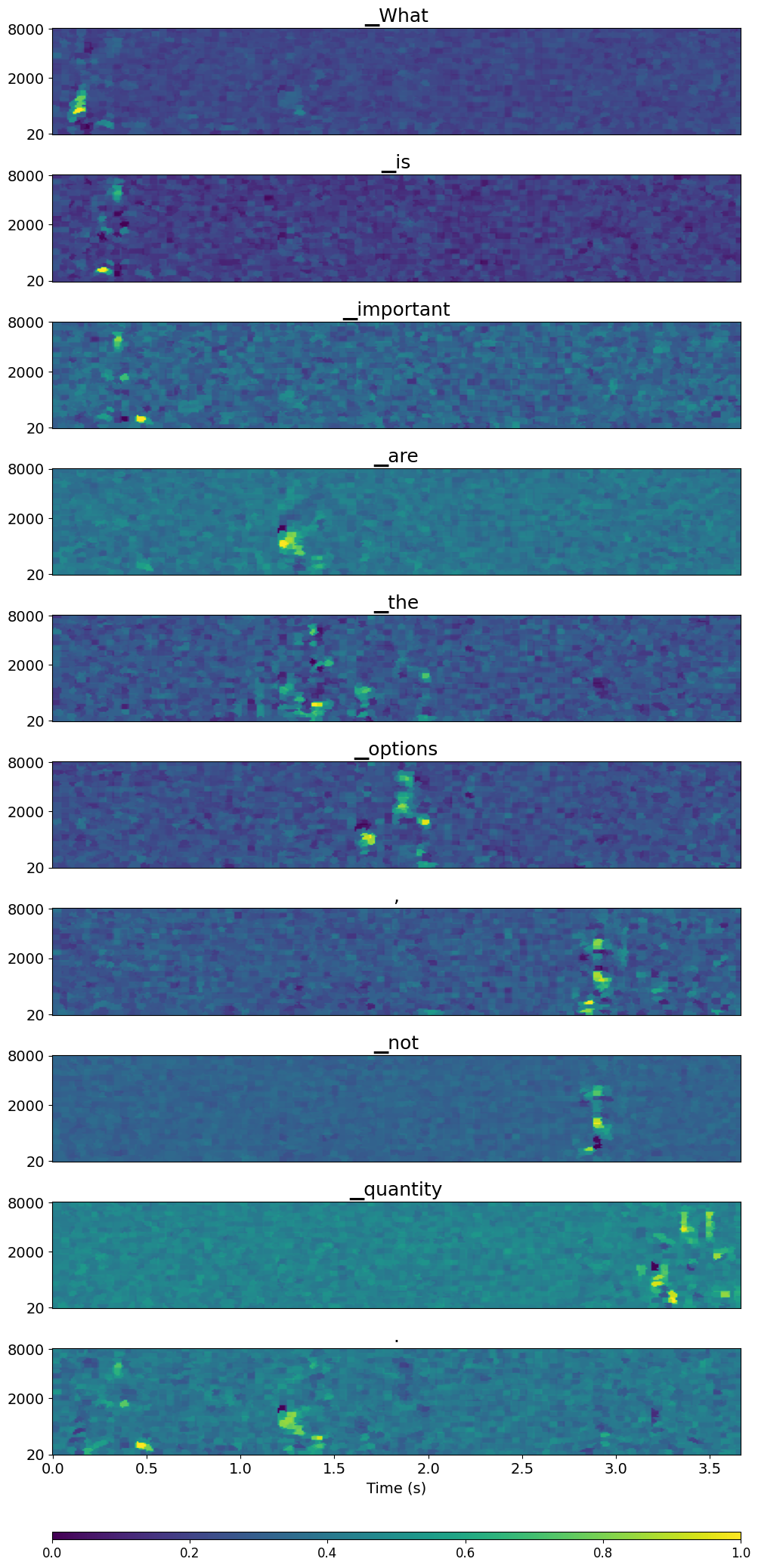}
    \caption{An example of \( \text{\SM{}}^X \) maps for the predicted sentence ``What is important are the options, not quantity''. The frequency axis is represented in Hertz on a logarithmic scale.}
    \label{fig:example_freq}
\end{figure}

\begin{table*}[!ht]
\centering
\footnotesize
\addtolength{\tabcolsep}{6.5pt}
\begin{tabular}{cl|cccccc:c|c}
\hline
\multicolumn{2}{c|}{\textbf{Aggregation}} &  &  &  &  &  &  &  \\
\cline{1-2}
\textit{frequency} & \multicolumn{1}{|c|}{\textit{time}} & \multirow{-2}{*}{$\ell=1$} & \multirow{-2}{*}{$\ell=2$} & \multirow{-2}{*}{$\ell=3$} & \multirow{-2}{*}{$\ell=4$} & \multirow{-2}{*}{$\ell=5$} & \multirow{-2}{*}{$\ell=6$} & \multirow{-2}{*}{{\textbf{$\ell$-AVG}}} & \multirow{-2}{*}{\textbf{ASR Del.$\uparrow$}} \\
\hline
\multicolumn{2}{c|}{2D avg} & \cellcolor[HTML]{FEFAE0}0.090 & \cellcolor[HTML]{FAF8DB}0.142 & \cellcolor[HTML]{E7EBC2}0.355 & \cellcolor[HTML]{E0E7B9}0.434 & \cellcolor[HTML]{DEE6B7}0.457 & \cellcolor[HTML]{DAE3B3}0.466 & \cellcolor[HTML]{DEE6B7}0.459 & 53.03 \\
\multicolumn{1}{l|}{1D max} & 1D avg & \cellcolor[HTML]{FCF9DE}0.115 & \cellcolor[HTML]{F7F6D7}0.176 & \cellcolor[HTML]{DFE6B8}0.441 & \cellcolor[HTML]{C0D091}0.534 & \cellcolor[HTML]{B6C883}0.560 & \cellcolor[HTML]{B4C781}0.565 & \cellcolor[HTML]{B4C681}0.565 & 55.18 \\
\multicolumn{2}{c|}{2D max} & \cellcolor[HTML]{FCF9DE}0.115 & \cellcolor[HTML]{F6F5D6}0.180 & \cellcolor[HTML]{DFE6B8}0.443 & \cellcolor[HTML]{BECE8E}0.540 & \cellcolor[HTML]{B1C47D}0.572 & \cellcolor[HTML]{ADC178}{\underline{0.582}} & \cellcolor[HTML]{B1C47D}\textbf{0.572} & \textbf{57.04} \\
\hline
\end{tabular}
\caption{Pearson $\rho$ correlation between layer-wise ($\ell$) cross-attention and the explanations, and the deletion scores (\textbf{ASR Del.}) for the different aggregation functions of the monolingual ASR model on English (\basem{}) on the dev set. The layer average (\textbf{$\ell$-AVG}) correlation is computed between the averaged cross-attention across layers ($\widebar{\text{\CA{}}}^{(\ell)}$) and the input explanations \SM{}\textsuperscript{$X$}. \textbf{Bold} indicates the highest result, \underline{underline} indicates the highest layer-wise correlation.}
    \label{tab:aggregation_fun}
\end{table*}

\subsection{Effect of Aggregation Functions on Input Explanations}
\label{subsec:aggregation-fun}

To properly obtain input-level explanations comparable with the dimensions of cross-attention scores (i.e., making \( \text{\SM{}}^X \in \mathbb{R}^{I \times T'} \)), we explore the effect of different aggregation strategies over the time and frequency dimensions. 


To compare and select the best aggregation strategy, we adopt the \textit{deletion} metric~\cite{nauta-etal-2023}, which quantifies the decline in prediction quality as the most relevant input frames--identified by the explanation--are progressively removed. Specifically, we adapt the implementation by \cite{fucci2025spes} for S2T tasks, replacing the top-ranked time frames in the input spectrogram \( \mathbf{X} \) with zero vectors in 5\% increments, based on the aggregated saliency map \( \text{\SM{}}^{X} \).
Since \( \text{\SM{}}^{X} \) operates on an aggregated time dimension \( T' \), which is smaller than the original time dimension \( T \) of \( \mathbf{X} \), we upsample \( T' \) to match \( T \) using nearest-neighbor interpolation.
Prediction quality is measured using the word error rate (WER), specifically the \texttt{wer\_max} scorer from the \href{https://github.com/hlt-mt/FBK-fairseq}{\texttt{\underline{SPES repository}}}. Lastly, we compute the area under the WER curve to quantify the faithfulness of each explanation method.

Table \ref{tab:aggregation_fun} reports the deletion scores and, for completeness, the Pearson $\rho$ correlations between the cross attention scores \CA{} and the saliency maps \SM{}\textsuperscript{$X$} for the representations aggregated following three strategies:
 \begin{itemize}
     \item \textbf{2D average pooling}, applied over the entire time-frequency plane to obtain its \textit{average} value and computed through \texttt{adaptive\_avg\_pool2d};
     \item \textbf{2-step pooling (1D maximum + 1D average)}, where max pooling is applied along the frequency axis, followed by averaging over time, and computed by applying \texttt{max\_pool1d}
     and \texttt{avg\_pool1d},
     respectively;
     \item \textbf{2D maximum pooling}, applied over the entire time-frequency plane to obtain its \textit{maximum} value and computed through \texttt{adaptive\_max\_pool2d}.
 \end{itemize}
The aggregation functions\footnote{All implemented in \href{https://docs.pytorch.org/docs/stable/nn.functional.html}{\texttt{\underline{pyTorch.nn.functional}}}.} were selected to contrast methods that either isolate the most relevant features (with maximum pooling) or represent their mean relevance (with average pooling). Similarly, the 2-step approach has been tried to first isolate relevance patterns in the frequency domain, a dimension that is not present in cross-attention representation, and then average across the time dimension to match the downsampled time resolution of the cross-attention scores.

Among the tested methods, we observe that the 2D maximum pooling aggregation (2D max) yields the best quality explanations, obtaining the highest deletion score, while the 2D average pooling (2D avg) is the worst, with the lowest deletion scores. About correlations, we notice that they follow the same trend of deletion scores, with 2D max yielding the best $\rho$. In particular, 2D avg consistently has the lowest correlations compared to the 2D max, particularly in the last layers (e.g., 0.457 vs. 0.572 at layer 5). Regarding the 2-step pooling approach, we not only observe an improved deletion score but also better correlation scores compared to 2D avg, especially from layer 3 onward, approaching the best performance with a layer-average correlation of 0.565. Nevertheless, the explanation quality is still lower compared to 2D max (55.18 vs. 57.04), which also achieves the highest correlations at nearly every layer, peaking at 0.582 in layer 6, and yielding the best overall correlation among the averaged cross-attention across layers (0.572). 
%
These results indicate that global averaging over time and frequency may obscure localized salient regions, and this is particularly impactful in the frequency dimension, where preserving saliency seems to play a crucial role.
This is due to the fact that key elements in the saliency maps are often well localized along the frequency axis. As shown in Figure~\ref{fig:example_freq}, for all tokens
saliency consistently concentrates in specific frequency bands. These bands are typically below 2000 Hz, where many resonant frequencies for speech are found \cite{stevens-2000}. As a result, smoothing operations such as 2D average pooling--or, to a lesser extent, the 2-step approach--tend to blur these concentrated regions, thereby diluting the saliency. This observation motivates our choice to adopt 2D max pooling in the main experiments.

\subsection{Quality Metrics for the Reported Models}
\label{subsec:model_quality}

The 
ASR quality
is evaluated with the WER metric using \href{https://pypi.org/project/jiwer}{\texttt{\underline{jiWER}}} and applying the \href{https://pypi.org/project/whisper-normalizer}{\texttt{\underline{Whisper text normalizer}}}.
The quality of the ST hypotheses is evaluated using COMET \cite{rei-etal-2020-comet} version 2.2.4, with the default model.
The quality of the explanations is obtained by measuring both 
\textit{deletion} and \textit{size} metrics
available in the SPES repository, using \texttt{wer\_max} as the scorer for ASR and \texttt{bleu} for ST, as described in \cite{fucci2025spes}.

\begin{table}[!ht]
\addtolength{\tabcolsep}{6pt}
\footnotesize
    \centering
    \begin{tabular}{c|cc|cc}
    \hline
       \multirow{2}{*}{\textbf{Model}}  & \multicolumn{2}{c|}{\textbf{WER $\downarrow$}} & \multicolumn{2}{c}{\textbf{COMET $\uparrow$}} \\
       \cline{2-5}
       & \textit{en} & \textit{it} & \textit{en-it} & \textit{it-en} \\
       \hline
        Whisper & 10.6 & \textbf{9.0} & - & 0.797 \\ 
        Seamless & 11.3 & \textbf{9.0} & 0.795 & \textbf{0.813} \\ 
        OWSM v3.1 & 11.9 & 17.0 & 0.634 & 0.559  \\ 
       \hdashline
        \texttt{base} & \textbf{9.5} & - & \multicolumn{2}{c}{-} \\
        \texttt{small} & 11.7 & 22.3 & 0.854 & 0.754 \\
        \texttt{large} & 11.1 & 21.7 & \textbf{0.862} & 0.765 \\
        \hline
        \multirow{2}{*}{\textbf{Model}} & \multicolumn{2}{c|}{\textbf{ASR Del. $\uparrow$}} & \multicolumn{2}{c}{\textbf{ST Del. $\downarrow$}}  \\
        \cline{2-5}
        & \textit{en} & \textit{it} & \textit{en-it} & \textit{it-en} \\
        \hline
        \texttt{base} &  91.3 & - & \multicolumn{2}{c}{-}  \\
        \texttt{small} & \textbf{92.6} & \textbf{97.0} & \textbf{2.4} & 2.4 \\
        \texttt{large} & 90.8 & \textbf{97.0} & 2.7 & \textbf{2.3}  \\
        \hline
        \multirow{2}{*}{\textbf{Model}} & \multicolumn{4}{c}{\textbf{Size $\downarrow$}} \\
        \cline{2-5}
        & \textit{en} & \textit{it} & \textit{en-it} & \textit{it-en} \\
        \hline
        \texttt{base} & 29.7 & - & \multicolumn{2}{c}{-} \\
        \texttt{small} & \textbf{29.4} & \textbf{28.2} & 30.0 & 29.4  \\
        \texttt{large} & 30.6 & 30.5 & \textbf{29.8} & \textbf{28.7} \\
         \hline
    \end{tabular}
    \caption{ASR and ST output quality (WER and COMET) and explanation quality (deletion and size) for all models analyzed in the paper on the EuroParl-ST test sets.}
    \label{tab:quality}
\end{table}

For comparison, in Table \ref{tab:quality}, we report results obtained from popular large-scale models, namely Whisper \cite{pmlr-v202-radford23a}, OWSM v3.1 \cite{peng24b_interspeech}, and SeamlessM4T \cite{barrault2023seamlessm4t}. Looking at the transcription/translation quality performance, we observe that both the monolingual \basem{} model and the multitask multilingual \smallm{} and \largem{} models are mostly able to achieve competitive results, even outperforming the well-known models in two cases (\textit{en} ASR for \basem{} and \textit{en-it} ST for \largem{}). While our models and OWSM v3.1 strive to be on par on \textit{it} with models with closed training data (Whisper and Seamless), they are able to close the gap on \textit{en}, most probably given a larger availability of public training data. Moreover, the highest performance of \basem{} on \textit{en} ASR compared to the \smallm{} and \largem{} can be attributed to both the specialization of the model and the presence of the EuroParl-ST training set in the training data. 
Moving to the explanation quality, we observe that both deletion and size scores are comparable across all three models analyzed in the paper and coherent with values obtained in the original SPES paper \cite{fucci2025spes} on different benchmarks and models. Overall, the deletion scores for ASR are close to the highest possible value (i.e., 100), especially on \textit{it}, where 97\% is achieved. Similarly, the deletion scores for ST are close to 0, indicating that the quality of explanations is very high. The size scores are all close, ranging between 28.2 and 30.6 among models, languages, and tasks, indicating a good compactness of the explanations.

\section{Results}
\subsection{
Does \CA{} Reflect Input-Output Dependencies?}
\label{subsec:input-results}


In this section, we compare \CA{} with input saliency maps \SM{}\textsuperscript{$X$}, which serve as an external reference for measuring input relevance.
Specifically, in \S\ref{subsubsec:mono_input_xai}, we analyze the 
\basem{} model across all levels of granularity. Then, in Section \ref{subsubsec:multitask_multilingual}, we extend the analysis to additional models (\smallm{} and \largem{}), languages (\textit{en} and \textit{it}), and tasks (ASR and ST).





\subsubsection{Head-wise and Layer-wise Correlations}
\label{subsubsec:mono_input_xai}


\begin{table*}[!ht]
\footnotesize
\centering
\addtolength{\tabcolsep}{6.5pt}
\begin{tabular}{c|cccccccc:c}
\bottomrule
{\footnotesize\textbf{Layer/Head}} & \multicolumn{1}{l}{$h=1$} & \multicolumn{1}{l}{$h=2$} & \multicolumn{1}{l}{$h=3$} & \multicolumn{1}{l}{$h=4$} & \multicolumn{1}{l}{$h=5$} & \multicolumn{1}{l}{$h=6$} & \multicolumn{1}{l}{$h=7$} & \multicolumn{1}{l|}{$h=8$} & \multicolumn{1}{l}{\textbf{$h$-AVG}} \\
\hline
$\ell=1$            & \cellcolor[HTML]{DAF4D7}0.076       & \cellcolor[HTML]{D0F4DE}-0.021      & \cellcolor[HTML]{DCF4D5}0.096       & \cellcolor[HTML]{D6F4DA}0.037       & \cellcolor[HTML]{D6F4D9}0.042       & \cellcolor[HTML]{D7F4D9}0.053       & \cellcolor[HTML]{D0F4DE}-0.020      & \cellcolor[HTML]{DCF4D5}0.094       & \cellcolor[HTML]{DEF4D4}0.111                   \\
$\ell=2$             & \cellcolor[HTML]{D3F4DC}0.013       & \cellcolor[HTML]{DFF4D3}0.122       & \cellcolor[HTML]{D6F4DA}0.039       & \cellcolor[HTML]{E4F4CF}0.171       & \cellcolor[HTML]{DFF4D3}0.123       & \cellcolor[HTML]{DBF4D6}0.089       & \cellcolor[HTML]{DFF4D3}0.119       & \cellcolor[HTML]{D6F4D9}0.041       & \cellcolor[HTML]{E5F4CE}0.178                   \\
$\ell=3$             & \cellcolor[HTML]{FDD7C0}0.455       & \cellcolor[HTML]{FCEEBD}0.404       & \cellcolor[HTML]{F7F5C1}0.348       & \cellcolor[HTML]{FBF5BE}0.386       & \cellcolor[HTML]{EEF5C7}0.263       & \cellcolor[HTML]{E0F4D2}0.136       & \cellcolor[HTML]{EDF5C9}0.248       & \cellcolor[HTML]{ECF5C9}0.246       & \cellcolor[HTML]{FCDDC0}0.443                   \\
$\ell=4$             & \cellcolor[HTML]{FCD8C0}0.452       & \cellcolor[HTML]{F7F5C1}0.344       & \cellcolor[HTML]{EAF5CA}0.227       & \cellcolor[HTML]{FCEEBE}0.405       & \cellcolor[HTML]{F4F5C3}0.314       & \cellcolor[HTML]{FDBDC3}0.512       & \cellcolor[HTML]{FCEABE}0.414       & \cellcolor[HTML]{FDC4C2}0.495       & \cellcolor[HTML]{FEADC5}0.546                   \\
$\ell=5$             & \cellcolor[HTML]{FDBEC3}0.508       & \cellcolor[HTML]{FDD3C1}0.463       & \cellcolor[HTML]{FDD2C1}0.466       & \cellcolor[HTML]{FEB8C4}0.521       & \cellcolor[HTML]{FDBAC4}0.518       & \cellcolor[HTML]{FAF5BF}0.374       & \cellcolor[HTML]{FDC1C3}0.502       & \cellcolor[HTML]{FDC9C2}0.485       & \cellcolor[HTML]{FE9EC7}0.578                   \\
$\ell=6$             & \cellcolor[HTML]{FBF5BE}0.377       & \cellcolor[HTML]{FCF3BD}0.394       & \cellcolor[HTML]{FDBFC3}0.508       & \cellcolor[HTML]{FCF6BD}0.386       & \cellcolor[HTML]{FDBAC4}0.517       & \cellcolor[HTML]{FAF5BF}0.371       & \cellcolor[HTML]{FCE5BF}0.424       & \cellcolor[HTML]{F5F5C3}0.322       & \cellcolor[HTML]{FF99C8}\textbf{\underline{0.588}}          \\
\hdashline
\textbf{$\ell$-AVG} & \cellcolor[HTML]{FEADC5}0.546       & \cellcolor[HTML]{FDD6C0}0.456       & \cellcolor[HTML]{FDD7C0}0.455       & \cellcolor[HTML]{FEA0C7}\underline{0.574}       & \cellcolor[HTML]{FEB5C4}0.529       & \cellcolor[HTML]{FEB0C5}0.541       & \cellcolor[HTML]{FEB6C4}0.525       & \cellcolor[HTML]{FDC2C3}0.500       & \cellcolor[HTML]{FE9FC7}0.577       \\
\hline
\end{tabular}
\caption{Pearson $\rho$ correlation between layer-wise ($\ell$) and head-wise ($h$) cross-attention and the explanations for the monolingual ASR model on English (\basem{}). The layer/head average (\textbf{$\ell$-/$h$-AVG}) correlation is computed between the averaged cross-attention across layers/head ($\widebar{\text{\CA{}}}^{(\ell)}$/$\widebar{\text{\CA{}}}_h$) and the input explanations \SM{}\textsuperscript{$X$}. \textbf{Bold} indicates the highest correlation, \underline{underline} indicates the highest layer-wise and head-wise correlation. Low to high values are \colorbox{green}{\textbf{\textsc{green}}}, to \colorbox{lightyellow}{\textbf{\textsc{yellow}}}, to \colorbox{pink}{\textbf{\textsc{pink}}}.}
    \label{tab:head_layer_wise}
\end{table*}

\begin{table*}[!ht]
\footnotesize
\addtolength{\tabcolsep}{0pt}
\begin{tabular}{c|c|l|ccccccclllll|c}
\bottomrule
\multicolumn{2}{c}{} & \multicolumn{14}{c}{\textbf{Target Language}} \\
\cline{3-16}
\multicolumn{2}{c}{} & \multicolumn{14}{c}{\textit{en}} \\
\cline{3-16}
\multicolumn{2}{c}{\multirow{-3}{*}{\textbf{Lang.}}} & \multicolumn{1}{c|}{\textbf{Model}} & {\scriptsize$\ell=1$} & {\scriptsize$\ell=2$} & {\scriptsize$\ell=3$} & {\scriptsize$\ell=4$} & {\scriptsize$\ell=5$} & {\scriptsize$\ell=6$} & {\scriptsize$\ell=7$} & {\scriptsize$\ell=8$} & {\scriptsize$\ell=9$} & {\scriptsize$\ell=10$} & {\scriptsize$\ell=11$} & \multicolumn{1}{c|}{{\scriptsize$\ell=12$}} & {\scriptsize$\ell$\textbf{-AVG}} \\
\hline
 &  & \smallm{} & \cellcolor[HTML]{FDFCDC}0.142 & \cellcolor[HTML]{D3F2D9}0.205 & \cellcolor[HTML]{3BCBCC}0.428 & \cellcolor[HTML]{1B96BA}{\textbf{\underline{0.639}}} & \cellcolor[HTML]{1A97BB}{\textbf{\underline{0.639}}} & \cellcolor[HTML]{159EBD}0.614 & \multicolumn{6}{c}{-} & \cellcolor[HTML]{1998BB}0.633 \\
 & \multirow{-2}{*}{\textit{en}} & \largem{} & \cellcolor[HTML]{F7FBDC}0.151 & \cellcolor[HTML]{F0F9DB}0.162 & \cellcolor[HTML]{CCF0D8}0.214 & \cellcolor[HTML]{85DED2}0.320 & \cellcolor[HTML]{9AE3D4}{0.289} & \cellcolor[HTML]{38CACB}0.434 & \cellcolor[HTML]{0EA8C0}0.581 & \cellcolor[HTML]{11A3BF}0.597 & \cellcolor[HTML]{149FBD}{\ultab 0.611} & \cellcolor[HTML]{11A3BF}0.597 & \cellcolor[HTML]{07B1C3}0.551 & \cellcolor[HTML]{0AAEC2}0.561 & \cellcolor[HTML]{169CBC}\textbf{0.621} \\
\cdashline{2-16}
 &  & \smallm{} & \cellcolor[HTML]{FDFCDC}0.147 & \cellcolor[HTML]{FDF4D4}0.193 & \cellcolor[HTML]{FDDDBB}0.327 & \cellcolor[HTML]{F7A18F}0.476 & \cellcolor[HTML]{F79F8D}{\underline{0.482}} & \cellcolor[HTML]{F8A692}0.465 & \multicolumn{6}{c}{-} & \cellcolor[HTML]{F69D8C}\textbf{0.485} \\
 & \multirow{-2}{*}{\textit{it}} & \largem{} & \cellcolor[HTML]{FDFBDB}0.151 & \cellcolor[HTML]{FDF9D9}0.164 & \cellcolor[HTML]{FDEFCE}0.223 & \cellcolor[HTML]{FDE2C0}0.300 & \cellcolor[HTML]{FDE4C3}0.285 & \cellcolor[HTML]{FCCAAC}0.383 & \cellcolor[HTML]{F8AC97}0.451 & \cellcolor[HTML]{F7A592}{\ultab 0.467} & \cellcolor[HTML]{F8A894}0.461 & \cellcolor[HTML]{F8A894}0.461 & \cellcolor[HTML]{FAB59E}0.430 & \cellcolor[HTML]{FAB59D}0.431 & \cellcolor[HTML]{F69A8A}\textbf{0.492} \\
 \cline{2-16}
 & \multicolumn{1}{l}{} & \multicolumn{14}{c}{\textit{it}} \\
\cline{2-16}
 &  & \smallm{} & \cellcolor[HTML]{FDF8D7}0.173 & \cellcolor[HTML]{FDF2D2}0.203 & \cellcolor[HTML]{FDDAB8}0.344 & \cellcolor[HTML]{F38279}0.547 & \cellcolor[HTML]{F28077}{\textbf{\underline{0.550}}} & \cellcolor[HTML]{F3857B}0.539 & \multicolumn{6}{c}{-} & \cellcolor[HTML]{F38178}0.549 \\
 & \multirow{-2}{*}{\textit{en}} & \largem{} & \cellcolor[HTML]{FDF8D8}0.168 & \cellcolor[HTML]{FDF7D7}0.176 & \cellcolor[HTML]{FDEDCC}0.235 & \cellcolor[HTML]{FDE2C0}0.300 & \cellcolor[HTML]{FDE1BF}0.306 & \cellcolor[HTML]{FBBDA3}0.413 & \cellcolor[HTML]{F59083}0.514 & \cellcolor[HTML]{F48B7F}0.526 & \cellcolor[HTML]{F48A7E}{\ultab 0.529} & \cellcolor[HTML]{F48A7E}{\ultab 0.529} & \cellcolor[HTML]{F59183}0.513 & \cellcolor[HTML]{F59082}0.516 & \cellcolor[HTML]{F28077}\textbf{0.551} \\
 \cdashline{2-16}
 &  & \smallm{} & \cellcolor[HTML]{FCFCDC}0.145 & \cellcolor[HTML]{D0F1D9}0.209 & \cellcolor[HTML]{60D4CF}0.374 & \cellcolor[HTML]{02B8C5}0.527 & \cellcolor[HTML]{03B6C5}{\underline{0.532}} & \cellcolor[HTML]{02B9C6}0.525 & \multicolumn{6}{c}{-} & \cellcolor[HTML]{05B4C4}\textbf{0.539} \\
\multirow{-9}{*}{\rotatebox{90}{\textbf{Source language}}} & \multirow{-2}{*}{\textit{it}} & \largem{} & \cellcolor[HTML]{ECF8DB}0.169 & \cellcolor[HTML]{F3FADC}0.157 & \cellcolor[HTML]{CCF0D8}0.215 & \cellcolor[HTML]{82DDD2}0.324 & \cellcolor[HTML]{95E2D3}0.297 & \cellcolor[HTML]{4ACECD}0.407 & \cellcolor[HTML]{0ABEC7}0.501 & \cellcolor[HTML]{08BEC7}0.503 & \cellcolor[HTML]{00BBC6}0.516 & \cellcolor[HTML]{00BBC6}{\ultab 0.518} & \cellcolor[HTML]{19C2C9}0.479 & \cellcolor[HTML]{17C1C8}0.482 & \cellcolor[HTML]{06B3C4}\textbf{0.544} \\
\hline
\end{tabular}
\caption{Person $\rho$ correlation between layer-wise cross-attention $\widebar{\text{\CA{}}}^{(\ell)}$ and the input explanations \SM{}\textsuperscript{$X$} for the multitask (ASR and ST) and multilingual (English and Italian) models (\smallm{} and \largem{}). \textbf{Bold} indicates the highest overall correlation, \underline{underline} indicates the highest correlation across layers. Low to high values are \colorbox{lightyellow}{\textbf{\textsc{yellow}}} to \colorbox{darkcyan}{\textbf{\textsc{aqua}}} for ASR, and to \colorbox{lightred}{\textbf{\textsc{red}}} for ST.}
    \label{tab:multilang_multitask}
\end{table*}

Table~\ref{tab:head_layer_wise} reports the 
correlations
for the monolingual English ASR model (\basem{}), considering cross-attention at the head level, layer level, and in aggregated form.
At the individual head level, correlations with saliency maps are generally low. This suggests that attention heads, when taken in isolation, only partially capture the model’s dependency on the input and often encode noisy or inconsistent relevance signals. However, not all heads are equal: some, especially in the upper layers (layers 4-6), exhibit relatively stronger correlations. Notably, \textbf{averaging across heads consistently outperforms selecting individual heads}, suggesting that, despite head-level sparsity and weak individual correlations, the collective information captured across heads reflects input relevance more effectively.
Moving from heads to layers, we find a clearer picture. Averaging attention 
across all heads within each layer substantially boosts correlation, with layer 6 standing out as the most aligned with the saliency maps. This is followed closely by layer 5 and the average across all layers, indicating that the \textbf{last layers exhibit the highest alignment with input relevance}. These results reinforce 
that deeper layers encode higher-level semantic or task-relevant features, a trend previously observed in Transformer-based models \cite{clark-etal-2019-bert}.
Interestingly, while averaging across heads improves alignment, averaging across both heads and layers does not yield the overall best result, even if values are close. This indicates that not all layers contribute equally and that indiscriminate aggregation can dilute the relevance signal.
Overall, the results show that appropriately selected and aggregated cross-attention scores exhibit only a \textit{moderate} to \textit{strong} correlation with input saliency maps, reaching values up to 0.588. This provides an initial indication of the 
limited explanatory power
of cross-attention weights, which we further examine under multilingual and multitask conditions in \S\ref{subsubsec:multitask_multilingual}.

\subsubsection{Multitask and Multilingual Correlations}
\label{subsubsec:multitask_multilingual}

To assess the impact of multilingual and multitask training on the correlation between cross-attention scores and saliency maps, we evaluate the \smallm{} and \largem{} models. Layer-wise results are shown in Table~\ref{tab:multilang_multitask}, while head-wise results are omitted due to the noisy behavior observed in \S\ref{subsubsec:mono_input_xai}.

Across all configurations, we observe that \textit{en} ASR yields the highest correlation values, outperforming even the monolingual \basem{} model (\S\ref{subsubsec:mono_input_xai}). This suggests that large-scale multilingual training enhances the alignment between cross-attention and saliency maps, likely due to the improved generalization capacity of the model. In contrast, \textit{en-it} ST shows a drop in correlation, which is expected given the increased complexity of ST compared to ASR.
When considering \textit{it} as the source language, we observe a similar pattern: ASR correlations are consistently higher than ST, yet remain below their \textit{en} counterparts. This discrepancy aligns with the data distribution in training, where \textit{en} accounts for 84\% of the data versus 16\% for \textit{it}, resulting in more robust representations for \textit{en}.

At the layer level, we find consistent evidence that the last decoder layers yield stronger correlations, reaffirming the trends observed in \S\ref{subsubsec:mono_input_xai}. 
The specific optimal layer varies with model size: layer 5 performs best in \smallm{}, while layers 8-10 achieve the highest correlations in \largem{}. 
Nevertheless, correlation values across the last
layers remain very close, suggesting that their cross-attention scores 
provide the most robust alignment with saliency maps across both tasks and languages. This trend is further supported by downstream application results, where the final layers have shown the best token-level performance \cite{papi-etal-2023-attention,papi23_interspeech,wang24ea_interspeech}.

Averaging attention scores across layers further improves the correlation with saliency maps in 
almost
all configurations. The only exceptions are \textit{en} and \textit{it} ASR in \smallm{}, where selective-layer extraction offers a marginal improvement (0.006 for English, 0.001 for Italian). 
Therefore, similarly to what we observed in \S\ref{subsubsec:mono_input_xai}, averaging attention across heads and layers consistently yields the best or near-best correlation with \textit{moderate} to \textit{strong} correlation with input saliency maps, even considering large-scale models trained in multitask and multilingual settings. 
Nonetheless, this alignment accounts for only 49-63\% of the total input relevance, indicating that \textcolor{darkpurple}{\textbf{cross-attention falls short of fully accounting for the S2T models' behavior}}. 
Since this limitation may stem from the phenomenon of context mixing, in \S\ref{subsec:context_mixing} we analyze the correlation between cross-attention and encoder output--representations that have already undergone a transformation by the encoder--to better isolate the true explanatory power of cross-attention.

  



\subsection{
What Is the Impact of Context Mixing?}
\label{subsec:context_mixing}

\begin{table*}[!t]
\footnotesize
\centering
\addtolength{\tabcolsep}{0pt}
\begin{tabular}{l|c|l|cccccccccccc|l}
\bottomrule
\multicolumn{2}{c}{} & \multicolumn{14}{c}{\textbf{Target Language}} \\
\cline{3-16}
\multicolumn{2}{c}{} & \multicolumn{14}{c}{\textit{en}} \\
\cline{3-16}
\multicolumn{2}{c}{\multirow{-3}{*}{\textbf{Lang.}}} & \textbf{Model} & {\scriptsize$\ell=1$} & {\scriptsize$\ell=2$} & {\scriptsize$\ell=3$} & {\scriptsize$\ell=4$} & {\scriptsize$\ell=5$} & {\scriptsize$\ell=6$} & {\scriptsize$\ell=7$} & {\scriptsize$\ell=8$} & {\scriptsize$\ell=9$} & {\scriptsize$\ell=10$} & {\scriptsize$\ell=11$} & {\scriptsize$\ell=12$} & {\scriptsize$\ell$\textbf{-AVG}} \\
\cline{1-16}
 &  & \basem{} & \cellcolor[HTML]{FDFCDC}0.105 & \cellcolor[HTML]{FDF5C5}0.179 & \cellcolor[HTML]{FFC541}0.602 & \cellcolor[HTML]{FFA022}0.708 & \cellcolor[HTML]{FF9316}{\ultab 0.745} & \cellcolor[HTML]{FFA526}0.693 & \multicolumn{6}{c}{-} & \cellcolor[HTML]{FF9014}\textbf{0.752} \\
 
 &  & \smallm{} & \cellcolor[HTML]{FDF9D3}0.137 & \cellcolor[HTML]{FDF0B8}0.220 & \cellcolor[HTML]{FED463}0.492 & \cellcolor[HTML]{FF9F20}{\ultab \textbf{0.712}} & \cellcolor[HTML]{FFA123}0.704 & \cellcolor[HTML]{FFB02F}0.664 & \multicolumn{6}{c}{-} & \cellcolor[HTML]{FFA122}0.706 \\
 
 & \multirow{-3}{*}{\cellcolor[HTML]{FFFFFF}\textit{en}} & \largem{} & \cellcolor[HTML]{FDF7CB}0.161 & \cellcolor[HTML]{FDF9D1}0.143 & \cellcolor[HTML]{FDF0B7}0.224 & \cellcolor[HTML]{FEE38F}0.352 & \cellcolor[HTML]{FDE69B}0.316 & \cellcolor[HTML]{FED568}0.476 & \cellcolor[HTML]{FFB433}0.651 & \cellcolor[HTML]{FFA525}0.695 & \cellcolor[HTML]{FFA425}0.697 & \cellcolor[HTML]{FFA324}{\ultab 0.699} & \cellcolor[HTML]{FFBC39}0.629 & \cellcolor[HTML]{FFB835}0.641 & \cellcolor[HTML]{FF9D1F}\textbf{0.718} \\
 
 \cdashline{2-16}
 
 &  & \smallm{} & \cellcolor[HTML]{BEE9FE}0.140 & \cellcolor[HTML]{B4E3F9}0.196 & \cellcolor[HTML]{96D0EA}0.362 & \cellcolor[HTML]{75C7E3}{\ultab 0.502} & \cellcolor[HTML]{77C7E4}0.495 & \cellcolor[HTML]{7EC8E5}0.467 & \multicolumn{6}{l}{} & \cellcolor[HTML]{70C6E3}\textbf{0.519} \\
 
 & \multirow{-2}{*}{\textit{it}} & \largem{} & \cellcolor[HTML]{C0EAFF}0.131 & \cellcolor[HTML]{C0EAFF}0.128 & \cellcolor[HTML]{B2E1F8}0.207 & \cellcolor[HTML]{A1D6F0}0.304 & \cellcolor[HTML]{A4D8F1}0.287 & \cellcolor[HTML]{8ECAE6}0.407 & \cellcolor[HTML]{72C7E3}0.511 & \cellcolor[HTML]{67C5E2}{\ultab 0.553} & \cellcolor[HTML]{6BC6E2}0.541 & \cellcolor[HTML]{69C5E2}0.545 & \cellcolor[HTML]{78C7E4}0.491 & \cellcolor[HTML]{73C7E3}0.509 & \cellcolor[HTML]{52C2DF}\textbf{0.633} \\
 
 \cline{2-16}
 & \multicolumn{1}{l}{} & \multicolumn{14}{c}{\textit{it}} \\
 \cline{2-16}
 
 &  & \smallm{} & \cellcolor[HTML]{B8E5FB}0.176 & \cellcolor[HTML]{ADDEF6}0.235 & \cellcolor[HTML]{86C9E5}0.439 & \cellcolor[HTML]{4CC2DE}{\ultab 0.655} & \cellcolor[HTML]{4DC2DF}0.650 & \cellcolor[HTML]{59C3E0}0.606 & \multicolumn{6}{c}{-} & \cellcolor[HTML]{4BC1DE}\textbf{0.659} \\
 
 & \multirow{-2}{*}{\textit{en}} & \largem{} & \cellcolor[HTML]{B6E4FA}0.184 & \cellcolor[HTML]{BBE7FD}0.160 & \cellcolor[HTML]{ACDDF5}0.241 & \cellcolor[HTML]{99D1EC}0.348 & \cellcolor[HTML]{97D0EB}0.357 & \cellcolor[HTML]{7AC8E4}0.484 & \cellcolor[HTML]{51C2DF}0.637 & \cellcolor[HTML]{49C1DE}0.667 & \cellcolor[HTML]{4AC1DE}0.661 & \cellcolor[HTML]{49C1DE}{\ultab 0.668} & \cellcolor[HTML]{54C3DF}0.625 & \cellcolor[HTML]{54C3DF}0.625 & \cellcolor[HTML]{44C0DD}\textbf{0.683} \\
 
 \cdashline{2-15}
 
 &  & \smallm{} & \cellcolor[HTML]{FDFAD4}0.133 & \cellcolor[HTML]{FDF3BF}0.199 & \cellcolor[HTML]{FEDE81}0.397 & \cellcolor[HTML]{FECB48}0.579 & \cellcolor[HTML]{FECA45}{\ultab 0.588} & \cellcolor[HTML]{FECD4F}0.555 & \multicolumn{6}{c}{-} & \cellcolor[HTML]{FFC843}\textbf{0.594} \\

\multirow{-10}{*}{\rotatebox{90}{\textbf{Source language}}} & \multirow{-2}{*}{\textit{it}} & \largem{} & \cellcolor[HTML]{FDF8CE}0.152 & \cellcolor[HTML]{FDFBD8}0.119 & \cellcolor[HTML]{FDF4C3}0.187 & \cellcolor[HTML]{FDE698}0.322 & \cellcolor[HTML]{FDE89E}0.303 & \cellcolor[HTML]{FEDB79}0.423 & \cellcolor[HTML]{FECD4E}0.560 & \cellcolor[HTML]{FFC440}0.605 & \cellcolor[HTML]{FFC440}0.606 & \cellcolor[HTML]{FFBF3C}{\ultab \textbf{0.619}} & \cellcolor[HTML]{FECB4A}0.573 & \cellcolor[HTML]{FECC4C}0.565 & \cellcolor[HTML]{FECB4A}0.573 \\
\hline
\end{tabular}
\caption{Person $\rho$ correlation between layer-wise cross-attention $\widebar{\text{\CA{}}}^{(\ell)}$ and the encoder output explanations \SM{}\textsuperscript{$H$} for all models (\basem{}, \smallm{} and \largem{}). \textbf{Bold} indicates the highest overall correlation, \underline{underline} indicates the highest correlation across layers. Low to high values for ASR are \colorbox{lightyellow}{\textbf{\textsc{yellow}}} to \colorbox[HTML]{FF9014}{\textbf{\textsc{orange}}}, and ST are \colorbox[HTML]{BFEAFF}{\textbf{\textsc{light cyan}}} to \colorbox[HTML]{44C0DD}{\textbf{\textsc{dark cyan}}}.}
\label{tab:encout}
\vspace{-0.7em}
\end{table*}

\begin{figure*}[ht]
    \centering
    \subfigure[\SM{}\textsuperscript{$X$}]{\includegraphics[width=0.35\linewidth]{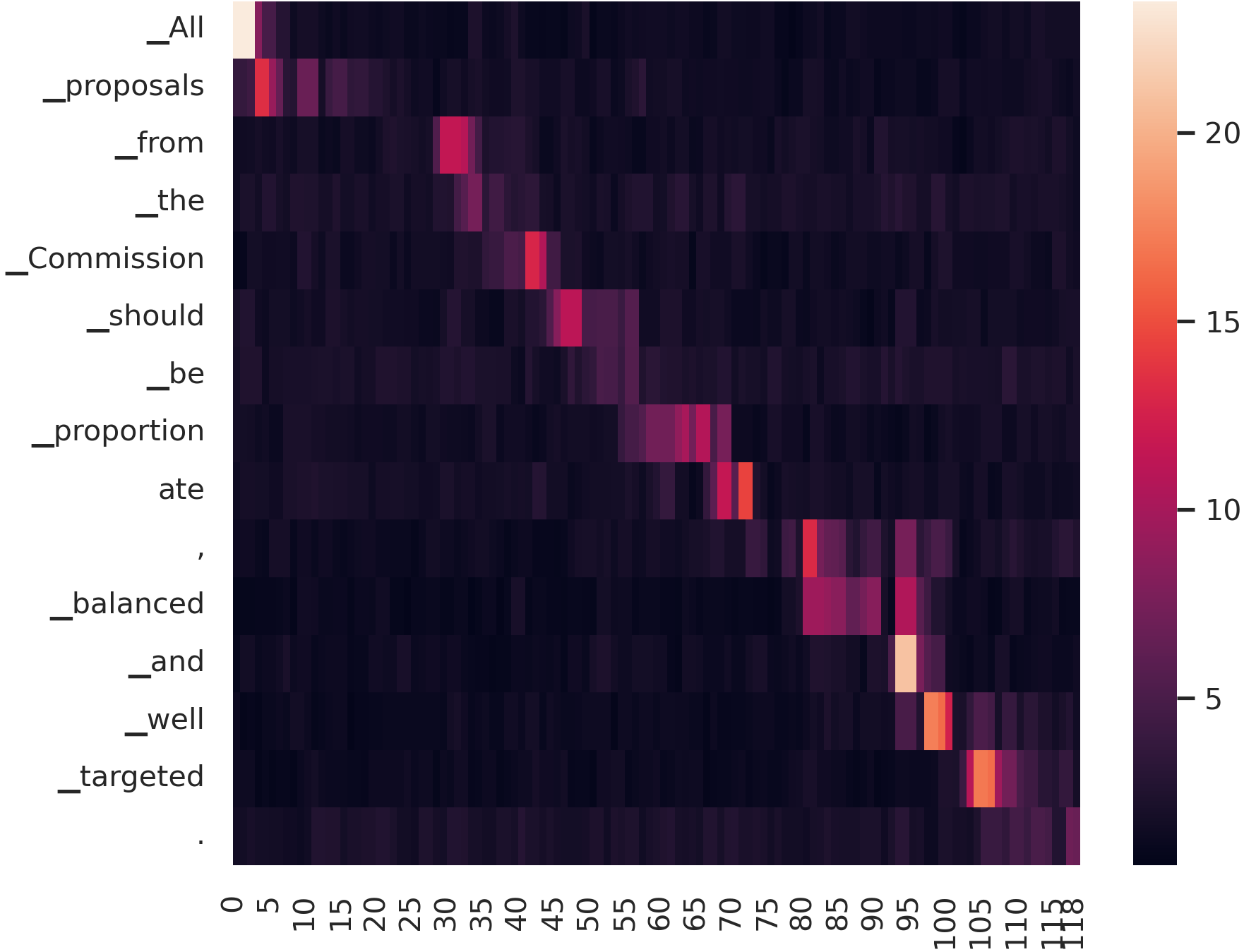}}
    \subfigure[\SM{}\textsuperscript{$H$}]{\includegraphics[width=0.295\linewidth]{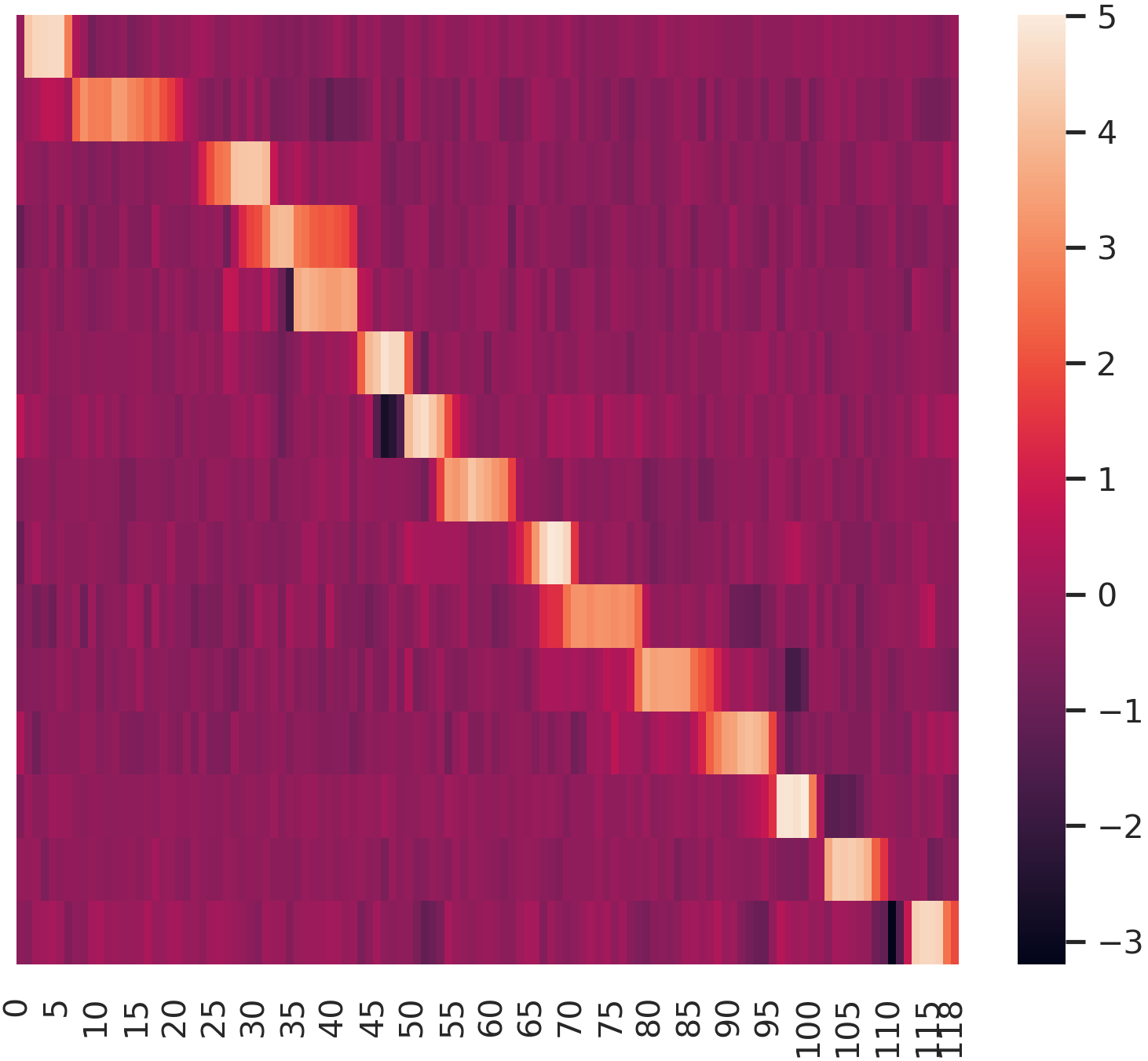}}
    \subfigure[\CA{}]{\includegraphics[width=0.295\linewidth]{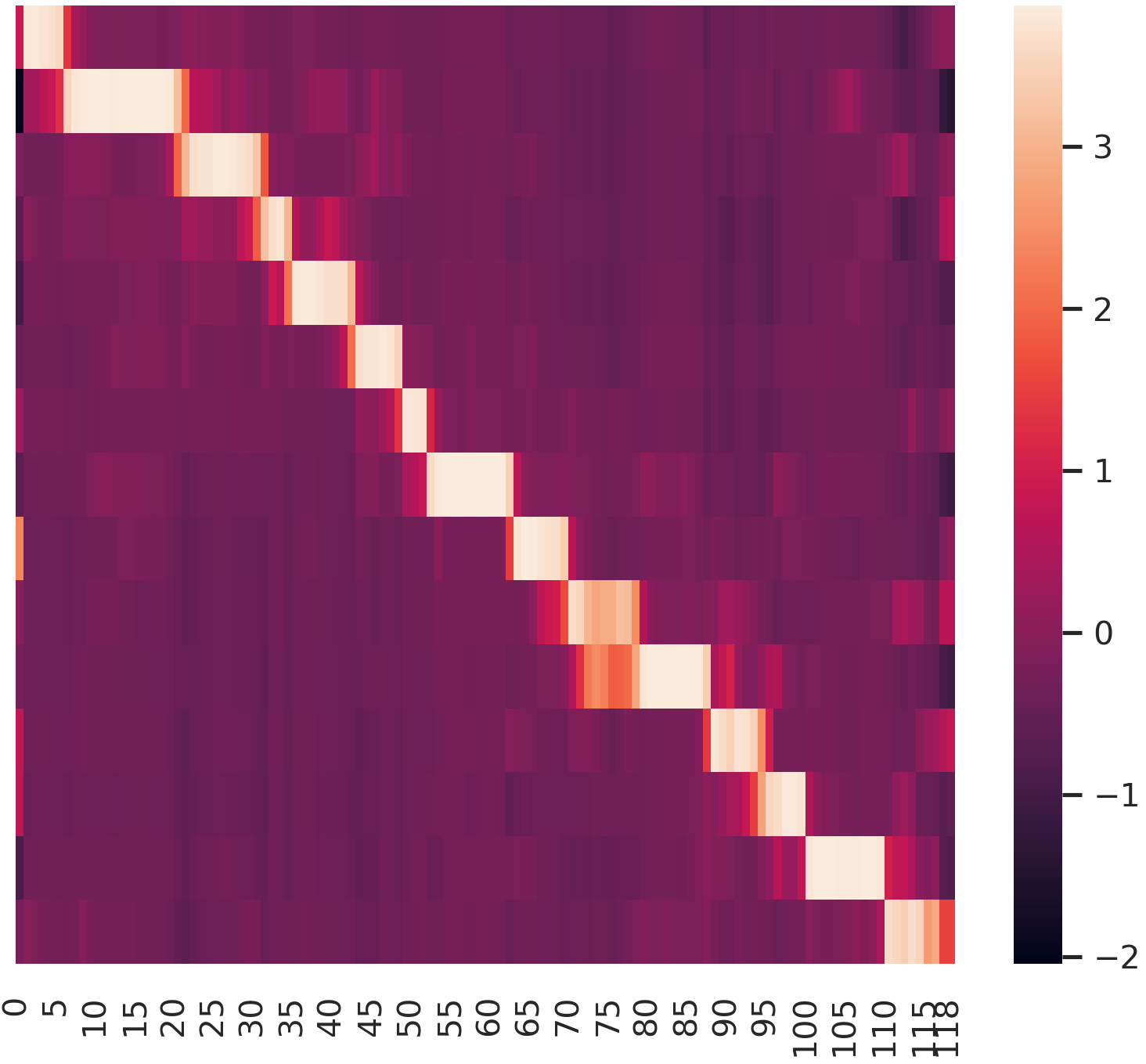}}
    \caption{Example extracted from the outputs produced by the \basem{} model on the dev set.
    }
    \label{fig:example}
    \vspace{-0.6em}
\end{figure*}

While \S\ref{subsec:input-results} focused on input relevance, we now investigate whether \CA{} 
aligns more closely with encoder output saliency maps. 
A higher correlation with encoder output representations would support the hypothesis that discrepancies between cross-attention and input saliency arise from context mixing, due to the reorganization of information within the encoder.
To this end,
we compare \CA{} with encoder output saliency maps \SM{}\textsuperscript{$H$}, which attribute relevance to the encoder hidden states for each output token (\S\ref{subsec:xai_s2t}). Layer-wise results for all models are presented in Table \ref{tab:encout}. 
%

Even when examining encoder output representations, we observe trends consistent with those identified in \S\ref{subsec:input-results}. Specifically, when averaged across decoder layers, cross-attention scores consistently provide the strongest or nearly optimal correlation with saliency maps, with the last decoder layers offering more representative explanations than the first ones across all models. As expected, correlation with encoder output representations consistently yields higher scores than those obtained from input representations, with absolute $\rho$ differences ranging from 0.03 to 0.18, quantifying the influence of context mixing effects to 6.6-16.7\%. 
The increased correlation is also visually evident in the example shown in Figure~\ref{fig:example}, where \CA{} aligns more closely with the relevance scores from \SM{}\textsuperscript{$H$} than with those from \SM{}\textsuperscript{$X$}. 
However, despite being unaffected by context mixing, the correlation between \CA{} and \SM{}\textsuperscript{$H$} remains limited--capturing only 52-75\% of the relevance.

\textcolor{darkpurple}{\textbf{This 
gap underscores the inherent limitations in relying solely on cross-attention as an explanation mechanism, reinforcing its role as an informative but \uline{incomplete} proxy for explainability in S2T models}}--not only for input-level saliency, but even at encoder-output level, where cross-attention directly operates.

\section{Discussion}
Our results show that, although \CA{} scores moderately correlate with aggregated  \(\text{\SM{}}^X\) (peaking around 0.45-0.55 in the best settings), they consistently fail to capture the full input relevance--even when context mixing effects are factored out. To directly assess explanation quality, we compute the deletion metric (see \S\ref{subsec:aggregation-fun}) on the \basem{} model, 
obtaining a score of 41.2 for \CA{}, compared to 52.9 for frequency-aggregated 
\(\text{\SM{}}^X\) and 91.3 for full-resolution maps. 
This gap indicates that \CA{} discards fine-grained time-frequency cues and produces weaker attributions, even under identical aggregation. 
Importantly, this trend is robust across both correlation and deletion scores, suggesting that the 
limitations of \CA{} 
reflect intrinsic constraints of attention as an explanatory signal rather than artifacts of the attribution reference.
These findings also carry implications for downstream tasks. In applications such as timestamp prediction, prior work often relies on attention from a single decoder layer or head \cite{zusag24_interspeech,wang24ea_interspeech,papi23_interspeech}. Our analysis suggests that averaging across layers and, especially, across heads 
yields attention patterns that more closely align with saliency behavior and may therefore improve the robustness of such methods.
Building on prior success of attention regularization in ASR (e.g., 
monotonic constraints
\cite{zhao20j_interspeech}), 
training-time strategies that explicitly encourage alignment between attention and attribution signals represent a promising direction to enhance both interpretability and task performance.
In summary, \textcolor{darkpurple}{\textbf{\underline{\CA{} should not be treated as a stand-alone XAI tool}}}. 
While it can provide lightweight, complementary cues, it 
is not 
a faithful proxy for input-output dependencies. Reframing \CA{} as an \textcolor{darkpurple}{\underline{\textbf{auxiliary signal}}} rather than a substitute for attribution-based explanations 
better grounds its use and points toward more principled approaches to explainability in S2T models.

\section{Conclusions}
We presented the first systematic analysis of cross-attention in S2T through the lens of explainable AI, comparing it to saliency maps across tasks, languages, and model scales. Cross-attention moderately to strongly aligns with saliency--especially when averaged across heads and layers--but captures only about half of the input relevance. Even when disentangling 
context mixing effects by analyzing encoder outputs, it explains just 52-75\% of saliency. This gap reveals intrinsic limits of cross-attention as an explanation mechanism: it offers informative cues but only a partial view of the factors driving S2T predictions.

\section{Limitations}
\label{app:limitations}

This work provides an in-depth analysis of cross-attention explainability in encoder-decoder S2T models. While it yields actionable insights, some limitations should be acknowledged. First, our experimental scope is restricted to ASR and ST. Although these tasks are central to S2T-based AI \cite{pmlr-v202-radford23a,barrault2023seamlessm4t}, we do not evaluate other downstream tasks such as spoken question answering or speech summarization, which may involve different dynamics in decoder attention. Second, our multilingual analysis is limited to English and Italian, due to the high computational cost of large-scale model training across a broader set of languages. Third, we focus on speech foundation models rather than SpeechLLMs,
a recent growing area of interest in S2T modeling \cite{gaido-etal-2024-speech}. 
Since our study focuses on evaluation, we aimed to avoid data contamination issues \cite{sainz-etal-2023-nlp}, a widespread concern in current SFMs and SpeechLLMs due to the lack of transparency about their training data. Therefore, we either retrained models from scratch or used fully open-science models with publicly documented training data.
Fourth, our analysis relies on SPES to compute reference explanations, acknowledging that, as an empirical method, it may introduce some margin of error. However, in the absence of a gold or human reference--which is unattainable in practice--we adopt SPES as a \textit{silver} reference, since it represents the state of the art in explainability for speech-to-text. We further validate this choice in \S\ref{subsec:model_quality}, showing that SPES achieves very high quality explanations (deletion scores $>$90 on ASR and $<$3 on ST), making it a more faithful option than less robust alternatives from the generic XAI field such as gradient norms \cite{JMLR:v22:20-1316}.


\section{Acknowledgments}
This paper has received funding from the European Union’s Horizon Europe programme grant agreement No. 101213369 (project DVPS).

\section{Generative AI Use Disclosure}
During the writing process, ChatGPT was employed exclusively to correct grammar in content authored by humans.

\bibliographystyle{IEEEtran}
\bibliography{mybib}

\end{document}